\documentclass[letterpaper]{article} % DO NOT CHANGE THIS
\usepackage{aaai25}  % DO NOT CHANGE THIS
\usepackage{times}  % DO NOT CHANGE THIS
\usepackage{helvet}  % DO NOT CHANGE THIS
\usepackage{courier}  % DO NOT CHANGE THIS
\usepackage[hyphens]{url}  % DO NOT CHANGE THIS
\usepackage{graphicx} % DO NOT CHANGE THIS
\usepackage{natbib}  % DO NOT CHANGE THIS AND DO NOT ADD ANY OPTIONS TO IT
\usepackage{caption} % DO NOT CHANGE THIS AND DO NOT ADD ANY OPTIONS TO IT
\usepackage{amsmath,amsfonts}
\usepackage{algorithm}
\usepackage{algorithmic}
\usepackage{bm}
\usepackage{ragged2e}
\usepackage{color}
\usepackage{multirow}
\usepackage[table]{xcolor}
\usepackage{threeparttable}
\usepackage{subcaption}
\usepackage{multicol}
\usepackage{cuted} % Provides the strip environment

\usepackage{newfloat}
\usepackage{listings}
\DeclareCaptionStyle{ruled}{labelfont=normalfont,labelsep=colon,strut=off} % DO NOT CHANGE THIS
\floatstyle{ruled}
\newfloat{listing}{tb}{lst}{}
\floatname{listing}{Listing}
\title{Precision-Enhanced Human-Object Contact Detection via Depth-Aware Perspective Interaction and Object Texture Restoration}
\author{
    Yuxiao Wang\textsuperscript{\rm 1}\equalcontrib, Wenpeng Neng\equalcontrib, Zhenao Wei\textsuperscript{\rm 1}, Yu Lei\textsuperscript{\rm 2}, Weiying Xue\textsuperscript{\rm 1}, Nan Zhuang\textsuperscript{\rm 3}, Yanwu Xu\textsuperscript{\rm 1}, Xinyu Jiang\textsuperscript{\rm 1}, Qi Liu\textsuperscript{\rm 1}\thanks{Corresponding author}
}
\affiliations{
    \textsuperscript{\rm 1}School of Future Technology, South China University of Technology\\
    \textsuperscript{\rm 2}School of Information Science \& Technology, Southwest Jiaotong University\\
    \textsuperscript{\rm 3}School of Software Technology, Zhejiang University

    ftwangyuxiao@mail.scut.edu.cn, nengwenpeng@gmail.com, drliuqi@scut.edu.cn
}

\usepackage{bibentry}
\begin{document}

\maketitle

\begin{abstract}
Human-object contact (HOT) is designed to accurately identify the areas where humans and objects come into contact. Current methods frequently fail to account for scenarios where objects are frequently blocking the view, resulting in inaccurate identification of contact areas.
To tackle this problem, we suggest using a perspective interaction HOT detector called PIHOT, which utilizes a depth map generation model to offer depth information of humans and objects related to the camera, thereby preventing false interaction detection. Furthermore, we use mask dilatation and object restoration techniques to restore the texture details in covered areas, improve the boundaries between objects, and enhance the perception of humans interacting with objects. Moreover, a spatial awareness perception is intended to concentrate on the characteristic features close to the points of contact. The experimental results show that the PIHOT algorithm achieves state-of-the-art performance on three benchmark datasets for HOT detection tasks. Compared to the most recent DHOT, our method enjoys an average improvement of 13\%, 27.5\%, 16\%, and 18.5\% on SC-Acc., C-Acc., mIoU, and wIoU metrics, respectively.
% Code is available at \url{https://drliuqi.github.io/}.
\end{abstract}


\begin{links}
    \link{Code}{https://drliuqi.github.io/}
    % \link{Datasets}{https://aaai.org/example/datasets}
    % \link{Extended version}{https://aaai.org/example/extended-version}
\end{links}

\section{Introduction}

\begin{figure*}[ht]
\centering
\includegraphics[width=\linewidth]{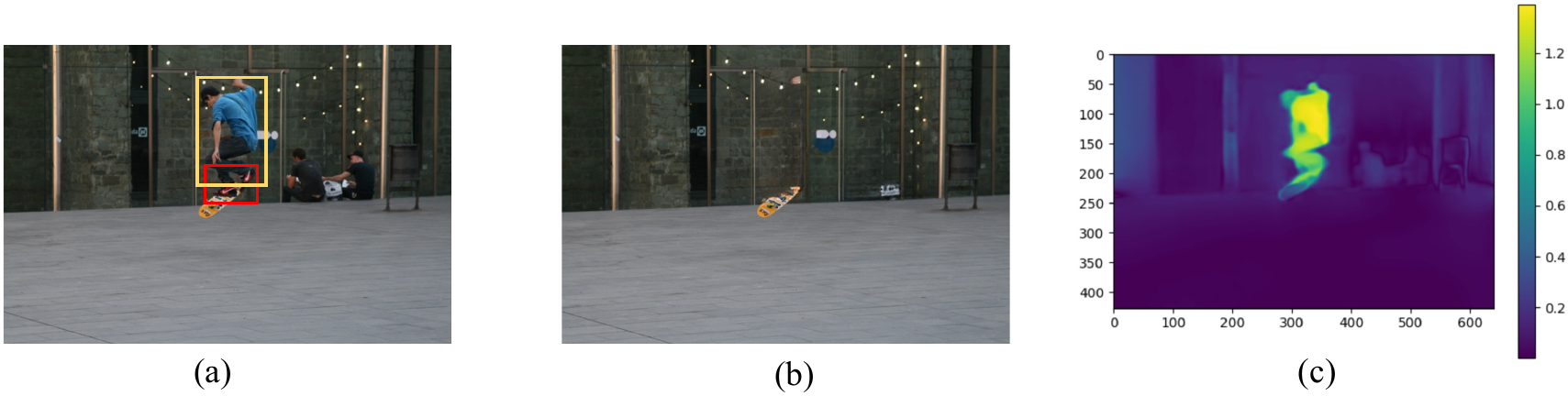}
\caption{In Figure (a), there are overlaps between the feet and the skateboard (highlighted by the red rectangle) as well as between the body and the wall (highlighted by the yellow rectangle) when viewed from a 2D plane perspective. In 3D space, only the feet are touching the skateboard. To gather information in 3D space, it is essential to reconstruct the objects, as illustrated in Figure (b), and then extract and compare their depth maps to determine the relative spatial positions in 3D space, as depicted in Figure (c). In Figure (c), it is evident that the person and the skateboard are situated on the same level in 3D space, whereas the person and the wall are on different levels. As a result, only the person and the skateboard are touching. The proposed method is founded on this principle for detecting human-object contact.}
\label{fig:index}
\end{figure*}

Human-object contact (HOT)~\cite{hot} refers to the process where humans come into proximity with and touch objects, a regular part of everyday life that helps us connect with the world around us. In Figure \ref{fig:index}, as a person skateboards, the area where their feet touch the skateboard creates a connection between the person and the skateboard. Research on HOT detection within the field of human-object interaction (HOI) focuses on accurately identifying the areas where humans and objects come into contact, allowing machines to precisely detect interactions between them. This technology has a wide range of potential applications and can be used in various areas such as human-computer interaction~\cite{wang2024ted}, virtual reality~\cite{illahi2023learning}, gesture recognition~\cite{cui2023hand}, among others, providing new opportunities in these fields.
\begin{figure}[ht]
\centering
\includegraphics[width=\linewidth]{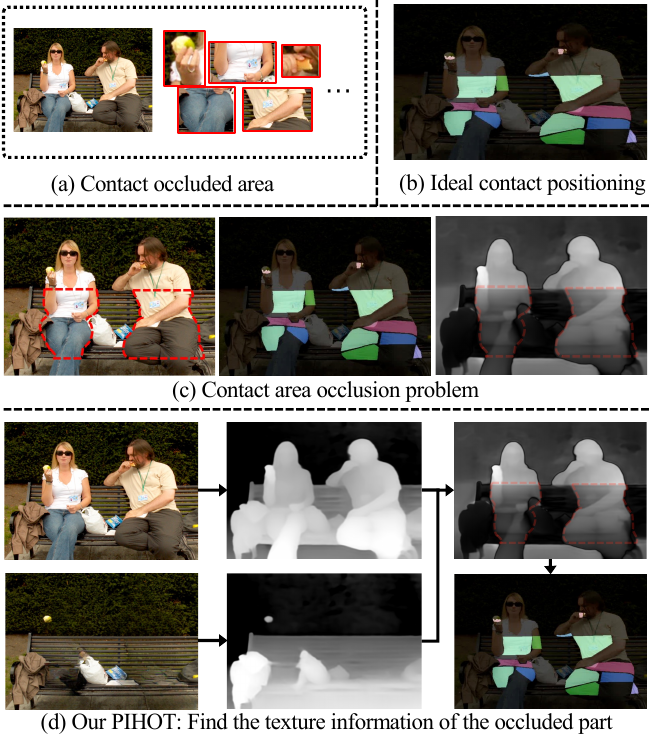}
\caption{Description of the issue with occlusion in the contact area. (a) The red rectangle on the right highlights the area where the contact is obscured in the input image. This phenomenon is natural when looking at a 2D image, which leads to reduced visibility of important edge details. As a result, the model finds it difficult to accurately identify contact areas. (b) Ideal contact positioning. (c) The problem of contact area. (d) Our proposed method aims to first fix the obstructed objects by restoring their original appearances. Next, it analyzes the depth features of the original and repaired images to gather their spatial positioning data. The depth map has successfully recovered the edge details of the obstructed objects.}
\label{fig:figure_1}
\end{figure}

At present, certain studies are mainly concentrated on particular areas such as hand-object contact~\cite{yang2024learning,shiota2024egocentric}, or they only take into account interactions between feet and the ground~\cite{rempe2021humor}. They rely on contact information as previous knowledge, which only identifies interactions between certain body parts and objects. Other studies focus on identifying interaction in particular settings~\cite{huang2022capturing, shimada2022hulc}, but the applicability of these methods is restricted and may not be relevant to more intricate situations. 

Current methods used for HOT fail to take into account the spatial relationship between humans and objects, resulting in possible misidentification of contact areas. In Figure \ref{fig:index}, in a two-dimensional view, the overlap of the human, skateboard, and wall may appear like the human is touching the wall, leading to a possible misinterpretation. Moreover, the current HOT method has not effectively utilized the interaction data between humans and objects. This approach~\cite{hot} simplifies HOT to a semantic segmentation task and does not take into account the frequent instances of occlusion by humans or objects.
Occlusion makes it difficult for them to identify the shape of the hidden area, which in turn impacts the accuracy of determining the boundary of contact. The bench's shape is lost because the human body covers it, as illustrated in Figure \ref{fig:figure_1}(a). Figure \ref{fig:figure_1}(b) displays the real contact labels, but the human completely obscures the segmentation boundary and object in Figure \ref{fig:figure_1}(c), presenting challenges in the HOT task. The intrinsic quality of the 2D image makes it challenging to deduce the meaning of the covered area based on the image's surroundings. The occlusion phenomenon limits the visibility of important edge information in the approach, resulting in inaccuracies when determining the contact area's location. Additionally, because the 2D image is just a projection from the 3D space, it lacks important spatial information in terms of depth. The spatial relationship is important evidence for determining if actual contact occurs in occlusion scenarios, as illustrated in Figure \ref{fig:figure_1}(d).

As a result, our proposal PIHOT utilizes depth and spatial awareness to investigate the relationship between humans and objects for interactive perspectives. PIHOT restores the shape of the original object by filling in the parts that are hidden and uses depth information to analyze the spatial relationship between humans and objects, leading to a deeper understanding of the scene. Specifically, PIHOT utilizes mask dilation to correct inaccuracies resulting from human errors in mask annotation and recovers object texture information lost in occluded areas through object inpainting to improve the reliability of inputs for the HOT task. By incorporating depth maps, spatial positions, and object information, PIHOT can create a more detailed feature representation near the contact area, allowing for a comprehensive understanding and accurate segmentation of the contact area.

In conclusion, this work's primary contributions include:

\begin{itemize}

\item [1)] A proposed technique using image inpainting aims to address the obscured issue of objects touching each other by restoring the shape contours and texture information in the occluded regions.

\item [2)] We aim to capture the distribution of depth information from objects in the camera's field of view to address the challenge of interpreting 3D information from 2D images. It leverages disparities in spatial distance between humans and objects to discern HOT interactions.

\item [3)] A spatial awareness method is suggested to improve the integration of the inpainting image with the depth map.

\item [4)] Extensive experiments demonstrate that the proposed method achieves state-of-the-art (SOTA) performance on three benchmark datasets.
\end{itemize}

\section{Related Work}

Like HOI, both HOI and HOT are necessary for identifying interactions between humans and objects. Nevertheless, there are some distinctions: HOT focuses mainly on precisely segmenting the particular contact areas between humans and objects, whereas HOI concentrates on identifying the interaction behaviors between humans and objects. Thus, HOT can be viewed as a more detailed breakdown of the HOI. To achieve this, the research will commence with the analysis of the HOI and explore current interaction models in more detail.

\begin{figure*}[ht]
\centering
\includegraphics[width=\linewidth]{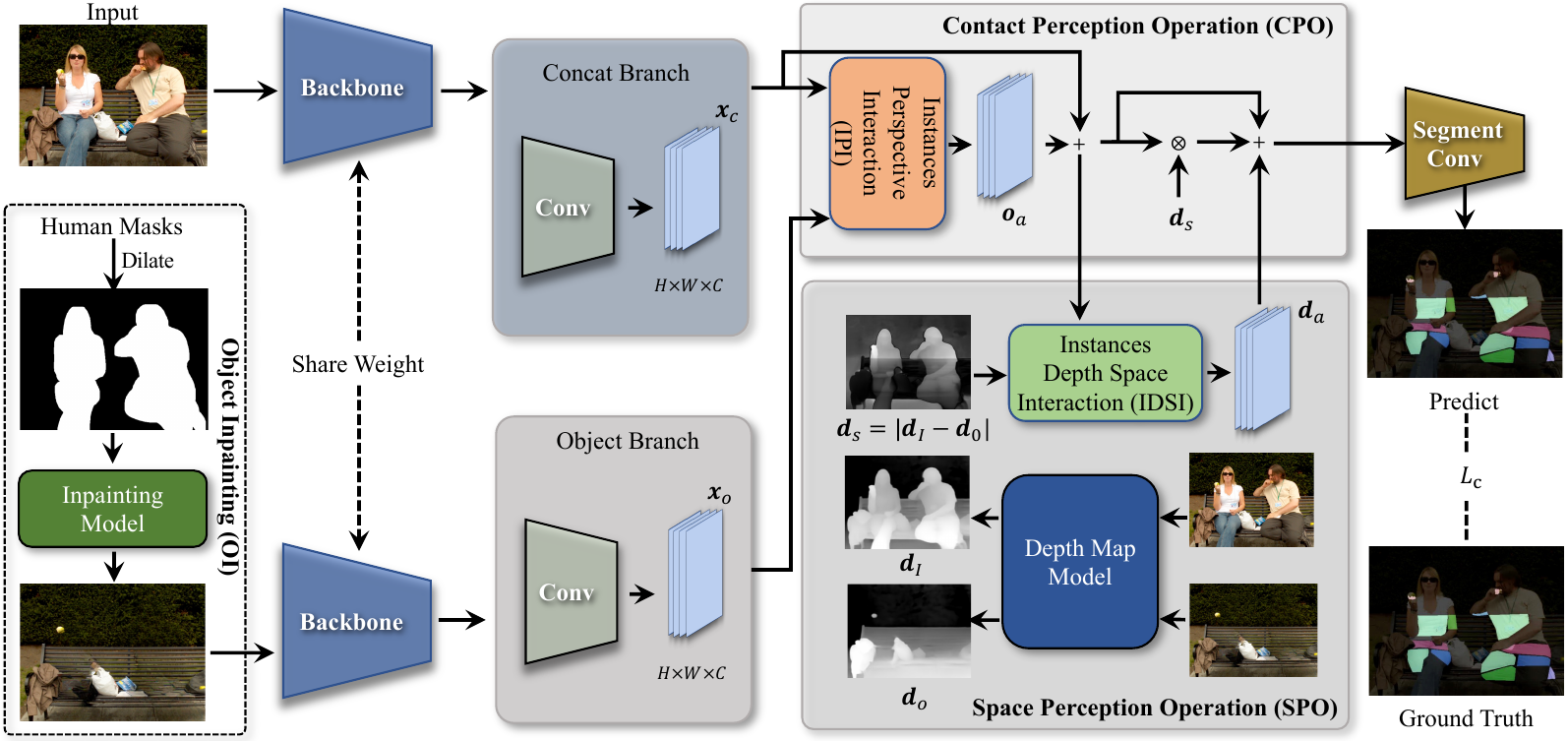}
\caption{The overall design of the PIHOT that is being proposed. PIHOT is primarily comprised of three main components: the object inpainting (OI) module, the contact perception operation (CPO) module, and the space perception operation (SPO) module. The OI module uses the human mask to restore the texture of occluded objects. The CPO module specifically targets the contact regions of the body parts in the initial image. The SPO module extracts depth maps from the original image and the repaired object image to encode how they are positioned with each other.}
\label{fig:network}
\end{figure*}

\subsection{Human-Object Interaction}

HOI detection mainly focuses on identifying how humans interact with objects in pictures or videos. Existing HOI detection models can be divided into twofolds: two-stage pipeline~\cite{liu2020consnet,tip-9552553,tip-9547056,tip-9489275} versus one-stage solution~\cite{zou2021end,liao2022gen,wang2024ted,tip-10328553,tip-10315051}. The first step in two-stage HOI detection involves using a pre-trained object detection model to identify both humans and objects, and then categorizing the interactions between each human and object. Separating object detection and interaction classification using these approaches results in decreased efficiency and accuracy~\cite{liao2022gen,wang2024ted}. As a result, one-step methods have been developed to tackle these challenges. Contrary to the two-stage approach, the one-stage model directly converts input images into HOI triples~\cite{zou2021end}.TED-Net~\cite{wang2024ted} featured a unique triple stream designed to concentrate on the context between humans and objects, enabling direct output of the positions of humans and objects as well as their interaction classification. Nevertheless, the HOI models mentioned above do not take into account if the interaction necessitates physical contact or specify which body parts of the human are being used in the interaction.

\subsection{Contact Modeling}
Several projects have been suggested for studying interactions between hands and objects~\cite{narasimhaswamy2020detecting,shan2020understanding,hasson2019learning,zhang2022fine,yang2024learning,shiota2024egocentric}. The study by~\cite{hasson2019learning} examined how to effectively apply constraints on reconstructing both hands and objects together. Tekin et al.~\cite{tekin2019h+} performed the joint estimation of 3D hand and object pose, along with the recognition of object and action categories. Nevertheless, these methods focus solely on the viewpoint of the first person and cannot determine the hand's contact status accurately.
The closely related solutions to this issue are the works by~\cite{narasimhaswamy2020detecting,shan2020understanding}. In the video-frame dataset, Shan et al.~\cite{shan2020understanding} labeled the positions of hands, sides of hands, states of contact, and locations of objects in contact. In addition, Narasimhaswamy et al.~\cite{narasimhaswamy2020detecting} used two attention mechanisms to understand details about hands and surrounding objects, which helped them determine the contact states of the hands.

Chen et al.~\cite{hot} presented a new challenge involving detecting human contact at the full-body level and developed the HOT dataset, which consists of 2D heatmaps showing contact areas and labels for corresponding human body parts. They take the contact regions between humans and objects into consideration and rely on a semantic segmentation model to segment the contact areas directly, which ignores the fact that contact positions are frequently occluded by objects. Consequently, the network has difficulty accurately segmenting the contact regions, resulting in blurred contact boundaries.

\section{Method}

We aim to utilize texture and spatial position information of both humans and objects to address the existing challenges of the HOT model, particularly its underperformance in complex scenes. To do that, we suggest using PIHOT to utilize depth image and repair images for detecting HOT when objects are obscured. Specifically, PIHOT initially retrieves object details using human masks. It uses the difference in depth between the original image and the repaired object image to calculate the physical distance between humans and objects. Moreover, the mechanisms of Instances Perspective Interaction (IPI) and Instances Depth Space Interaction (IDSI) are intended to help the network prioritize occluded object information, improve the contact boundaries between humans and objects, and offer depth information of humans and objects related to the camera within the scene. The network framework is illustrated in Figure \ref{fig:network}. 

Given an image $\bm{I}$, $\bm{x}$ is extracted after passing through the backbone network. Subsequently, $\bm{x}$ is fed into the contact branch to extract contact attention feature $\bm{x}_c$. Meanwhile, the repaired object image $\bm{I}_o$ is obtained by the object inpainting module, and feature $\bm{x}_o$ is extracted through the backbone network. The parameters of these two backbone networks are shared. The space perception operation (SPO) module extracts depth maps from the original image $\bm{I}$ and the repaired object image $\bm{I}_o$, resulting in $\bm{d}_i$ and $\bm{d}_o$, respectively. By computing the difference between $\bm{d}_i$ and $\bm{d}_o$, the spatial relative position feature map $\bm{d}_s$ between humans and objects is obtained, which is then passed into the proposed IDSI mechanism to get the output $\bm{d}_a$. The contact perception operation (CPO) module combines $\bm{x}_c$, $\bm{x}_o$, $\bm{d}_s$, and $\bm{d}_a$ for final contact feature extraction, thereby generating HOT segmentation map.

\begin{figure}[h]

\centering %表示居中
\includegraphics[width=\linewidth]{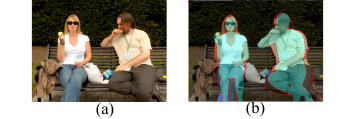}

% \vspace{-0.3cm}
\caption{Description of human-annotated masks.}

\label{fig:occlusions}

% \vspace{-0.2cm}
\end{figure}

\textbf{Object Inpainting Module}. During OI, the main objective is to eliminate humans from the background by utilizing mask information. The dataset labels typically include this information about the masks. However, it is worth mentioning that manually annotated mask information may contain some inaccuracies, such as labeling errors. Errors in the human body mask annotations are present for each image in the dataset~\cite{hot}. As shown in Fig. \ref{fig:occlusions}(b), the original annotations (the baby blue area) do not fully cover the human body (the red dashed area). Using a dilatation rule can help extend the human body annotations to correct any small errors and ensure the entire object is captured more accurately. As a result, the mask is resized through a dilation operation to expand the area to encompass all crucial information about humans. The dilatation is specifically determined by the following 
formulas:
\begin{equation}
    \bm{k}=\begin{bmatrix}  
  1 & \cdots & 1 \\  
  \vdots & \ddots & \vdots \\  
  1 & \cdots & 1  
\end{bmatrix}^{N\times N},
\end{equation}
\begin{equation}
    \bm{m}=\texttt{conv2d}(\bm{m},\bm{k}),
\end{equation}
where $N$ represents the row or column numbers of the matrix, \texttt{conv2d} denotes the 2D convolution operation, $\bm{m}$ represents the human mask, and $\bm{k}$ is the convolution kernel of size $N\times N$ with all elements being 1. The next step involves constructing and initializing the OI model with the parameters from the LaMa model~\cite{suvorov2022resolution}. This step removes any human presence from the image, leaving only object information.

\textbf{Instances Perspective Interaction Mechanism}. Merging the characteristics of $\bm{x}_c$ and $\bm{x_o}$ in the channel is a basic method of fusion, but it does not effectively leverage object information to restrict the features obtained from the original image. As a result, we develop the IPI mechanism. The structure diagram can be found in the supporting materials. Specifically, the feature $\bm{x}_o$ undergoes expansion into a 2D matrix through a convolutional layer, leading to a matrix $\bm{Q}$. Next, the characteristics of $\bm{x}_c$ are inputted into two separate convolutional layers to generate matrices $\bm{K}$ and $\bm{V}$. The importance of each position, or the weights, can be obtained by multiplying matrix $\bm{Q}$ by the transpose of matrix $\bm{K}$. Lastly, when these weights are multiplied by matrix $\bm{V}$, the resulting feature matrix is $\bm{o}_a$.

\begin{align}
    \bm{Q} = \texttt{E}(\texttt{conv2d}(\bm{x}_b )), \nonumber \\
    \bm{K} = \texttt{E}(\texttt{conv2d}(\bm{x}_c )), \nonumber \\
    \bm{V} = \texttt{E}(\texttt{conv2d}(\bm{x}_c)),
\end{align}
\begin{align} \label{eq:feature_attention}
\bm{o}_a & = \texttt{Feature\_Attention}(\bm{Q},\bm{K},\bm{V}) \nonumber \\
& = softmax(\frac{\bm{Q}\bm{K}^T}{\sqrt{d}}\bm{V}),
\end{align}
where $\texttt{E}(\cdot)$ represents the expand operation. $d$ denotes the number of columns in the matrix $\bm{Q}$, $\bm{K}$, or $\bm{V}$. The weights for each \texttt{conv2d} operation are randomly initialized.

\textbf{Instances Depth Space Interaction}. We have developed a mechanism called IDSI to make the most of the feature information provided by the depth map. It aims to utilize the proximity between humans and objects to reach the point of contact.
The structure of IDSI can be found in the supporting materials.
Specifically, we subtract the depth feature map of the object $\bm{d}_o$ from the depth feature map of the original image $\bm{d}_i$ to extract the feature information related to relative spatial positions $\bm{d}_s$. It is important to apply normalization to ensure that the depth map information falls within a specific range, which is accomplished by:
\begin{equation} \label{eq:d_s_1}
    \bm{d}_s=\left | \bm{d}_i-\bm{d}_o \right | ,
\end{equation}
\begin{equation} \label{eq:d_s_2}
    \bm{d}_s=\frac{\bm{d}_s-\texttt{min}(\bm{d}_s)}{\texttt{max}(\bm{d}_s)-\texttt{min}(\bm{d}_s)},
\end{equation}
where $\left| \cdot \right|$ represents the absolute value operator, and \texttt{max} and \texttt{min} indicate taking the maximum and minimum values of the vector, respectively. This step enforces the relative spatial distance $\bm{d}_s$ to be positive and maintains the accuracy of the depth map data. In an ideal situation, by subtracting the depth maps ($\bm{d}_s$), we would be left with only the human and the occluded object. 

Nevertheless, the repaired image shows a noticeable difference in depth maps compared to the original image generated by the model, with the body appearing brighter, the background and object parts darker, and occlusions having intermediate brightness, as depicted in the right subfigure of Figure \ref{fig:figure_1}(c). Therefore, we are motivated to separate the occlusive and occlusion-free parts due to the increasing visibility of the boundary of occlusion.

Next, the output $\bm{o}_a$ of the IPI module is fed to the convolutional layer, and then combined with $\bm{d}_s$ to input the feature attention module, via:
\begin{align}
    d_a=\texttt{Feature\_Attention}(\bm{S}), \nonumber \\
    \bm{S} = d_s,\texttt{E}(\texttt{conv2d}(o_a )),\texttt{E}(\texttt{conv2d}(o_a))
\end{align}
where \texttt{Feature\_Attention} is identical to the Eq. \ref{eq:feature_attention}. The IDSI mechanism enables the model to incorporate depth information more thoroughly. It can enhance the understanding of spatial relationships between humans and objects, as well as improve the accuracy and reliability of human object tracking.

\textbf{Contact Perception Operation Module}. This module is intended for receiving the contact feature $\bm{x}_c$, the IPI feature $\bm{o}_a$, and the depth feature $\bm{d}_a$ to be utilized in the final segmentation feature extraction process. It specifically adheres to the following equations:
\begin{equation} \label{eq:alpha_beta}
    \bm{x}=(\bm{x}_c+\alpha \times \bm{o}_a)\times \bm{d}_s+(\bm{x}_c+\alpha \times \bm{o}_a)+\beta\times \bm{d}_a,
\end{equation}
where $\bm{d}_s$ represents the normalized result of subtracting the object-repaired depth map $\bm{d}_o$ from the original image depth map $\bm{d}_i$. Values of $\bm{d}_s$ are expected to be near 0 for matching areas in the original and object-repaired images, and above 0 for areas that do not match. Therefore, by using $(\bm{x}_c+\alpha \times \bm{o}_a)\times \bm{d}_s+(\bm{x}_c+\alpha \times \bm{o}_a)$, the feature $(\bm{x}_c+\alpha \times \bm{o}_a)$ are highlighted for the differing parts between the original and object-repaired images. Subsequently, the result is combined with the IDSI $\bm{d}_a$ to produce the comprehensive output. $\alpha$ and $\beta$ are weight hyperparameters. $\bm{x}$ is forwarded to the segment conv module to generate the features for the HOT segmentation map. This module consists of three sets of $1\times 1$ convolutions followed by batch normalization and ReLU activation functions. Finally, the output passes through a $1\times 1$ convolutional layer with $18$ output channels to obtain the final segmentation probability map. These $18$ channels represent different HOT categories, with a total of 17 classes excluding the background. The detailed introduction of the specific HOT categories can be found in the supporting materials.

\textbf{Loss.} Given image $\bm{I}$, PIHOT will predict a probability map $\hat{\bm{y}} \in \mathbb{R}^{H\times W\times C_{y}}$, where $C_y$ represents the number of HOT classes, and H and W are the height and width of the probability map, respectively. $\hat{\bm{y}}$ is compared with the ground truth $\bm{y}$ to compute the cross-entropy loss. Specifically, each pixel of $\hat{\bm{y}}$ denotes a classification example, and the cross-entropy loss is computed for each corresponding pixel in $\bm{y}$, via:
\begin{equation}
    {L}_{c}=-\displaystyle\sum_{i=1}^{H}\displaystyle\sum_{j=1}^{W}\displaystyle\sum_{k=1}^{C_{\bm{y}}}\bm{y}^k_{i,j}\log{\hat{\bm{y}}^k_{i,j}}+(1-\bm{y}^k_{i,j})\log(1-\hat{\bm{y}}^k_{i,j})
\end{equation}

\section{Results}

\begin{table*}[ht]
\centering
\begin{tabular}{l|cccc|cccc}
\hline
\multirow{2}{*}{Model} & \multicolumn{4}{c}{HOT-Annotated} & \multicolumn{4}{c}{HOT-Generated}   \\ \cline{2-9} 
                       & SC-Acc.  & C-Acc.  & mIoU    & wIoU & SC-Acc.  & C-Acc.  & mIoU    & wIoU  \\ \hline
ResNet+UperNet~\cite{xiao2018unified}         & 35.1     & 62.6   & 0.195   & 0.227 & 21.1     & 42.7   & 0.080  & 0.116 \\
ResNet+PPM~\cite{zhao2017pyramid}             & 34.6     & 61.1   & 0.201   & 0.233 & 21.2     & 41.1   & 0.075  & 0.119  \\
DHOT$_\text{wo/att}$~\cite{hot}                    & 24.1     & 42.8   & 0.148   & 0.187 & 12.0     & 24.6   & 0.051  & 0.099  \\
DHOT$_\text{pure\_att}$~\cite{hot}                    & 33.8     & 58.4   & 0.189   & 0.237 & 20.3     & 40.1   & 0.077  & 0.113  \\
DHOT$_\text{Full}$~\cite{hot}                    & 40.7     & 70.7   & 0.215   & 0.260 & 30.4     & 54.3   & 0.139  & 0.167 \\
\rowcolor[gray]{0.9}PIHOT                   & \begin{tabular}[c]{@{}c@{}}\textbf{45.3}\\(\textcolor{red}{+11\%})\end{tabular}    & \begin{tabular}[c]{@{}c@{}}\textbf{80.7}\\(\textcolor{red}{+14\%})\end{tabular}  & \begin{tabular}[c]{@{}c@{}}\textbf{0.236}\\(\textcolor{red}{+10\%})\end{tabular}  & \begin{tabular}[c]{@{}c@{}}\textbf{0.286}\\(\textcolor{red}{+10\%})\end{tabular} & \begin{tabular}[c]{@{}c@{}}\textbf{34.9}\\(\textcolor{red}{+15\%})\end{tabular}       & \begin{tabular}[c]{@{}c@{}}\textbf{76.3}\\(\textcolor{red}{+41\%})\end{tabular}     & \begin{tabular}[c]{@{}c@{}}\textbf{0.169}\\(\textcolor{red}{+22\%})\end{tabular}     & \begin{tabular}[c]{@{}c@{}}\textbf{0.212}\\(\textcolor{red}{+27\%})\end{tabular}\\ \hline
\end{tabular}
\caption{Assessment of contact detection accuracy using the HOT-Annotated dataset.}
\label{tab:results-annotated}
\end{table*}

\begin{figure}[ht]
    \centering
    \begin{subfigure}{\linewidth}
        \includegraphics[width=\linewidth]{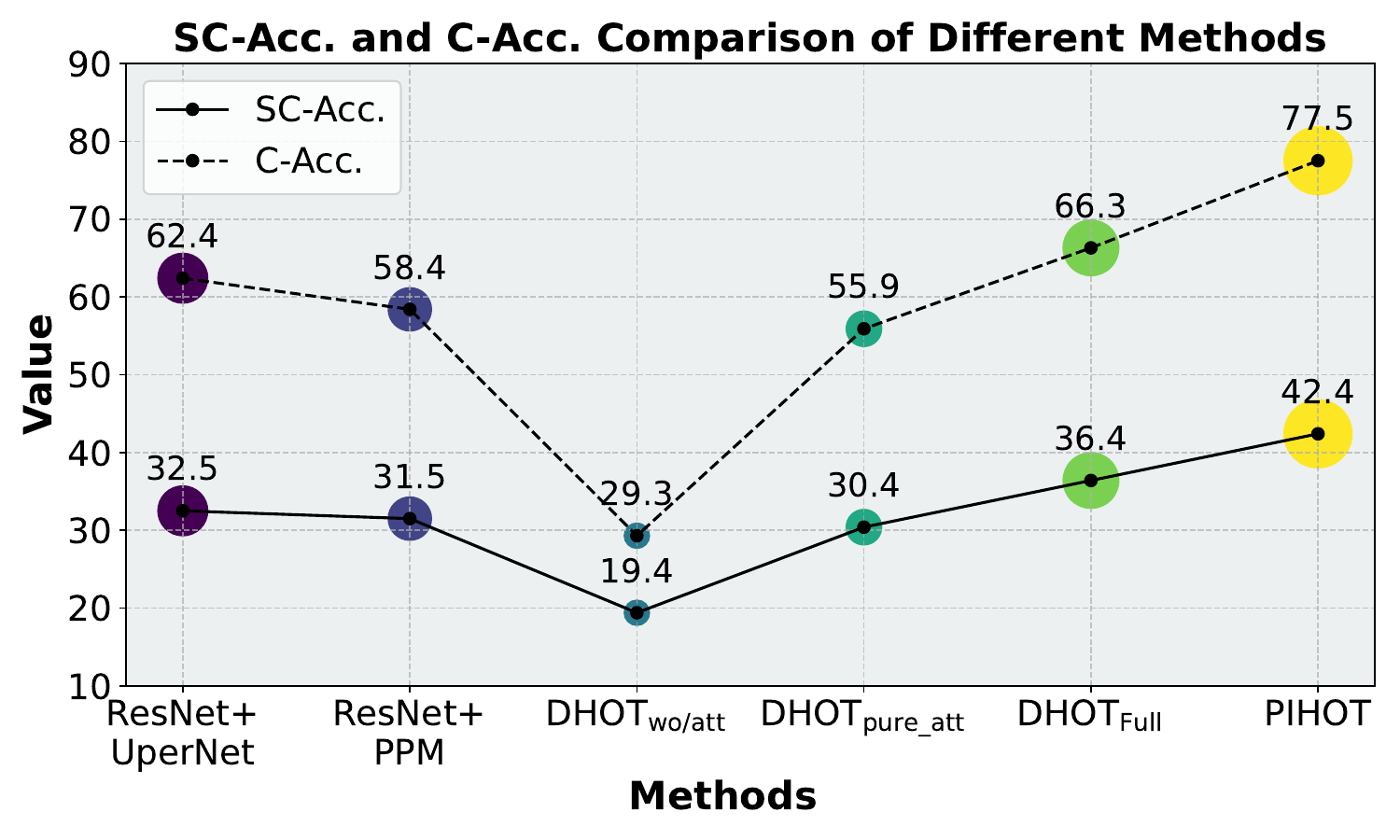}
        \caption{}
    \end{subfigure}
    \begin{subfigure}{\linewidth}
        \includegraphics[width=\linewidth]{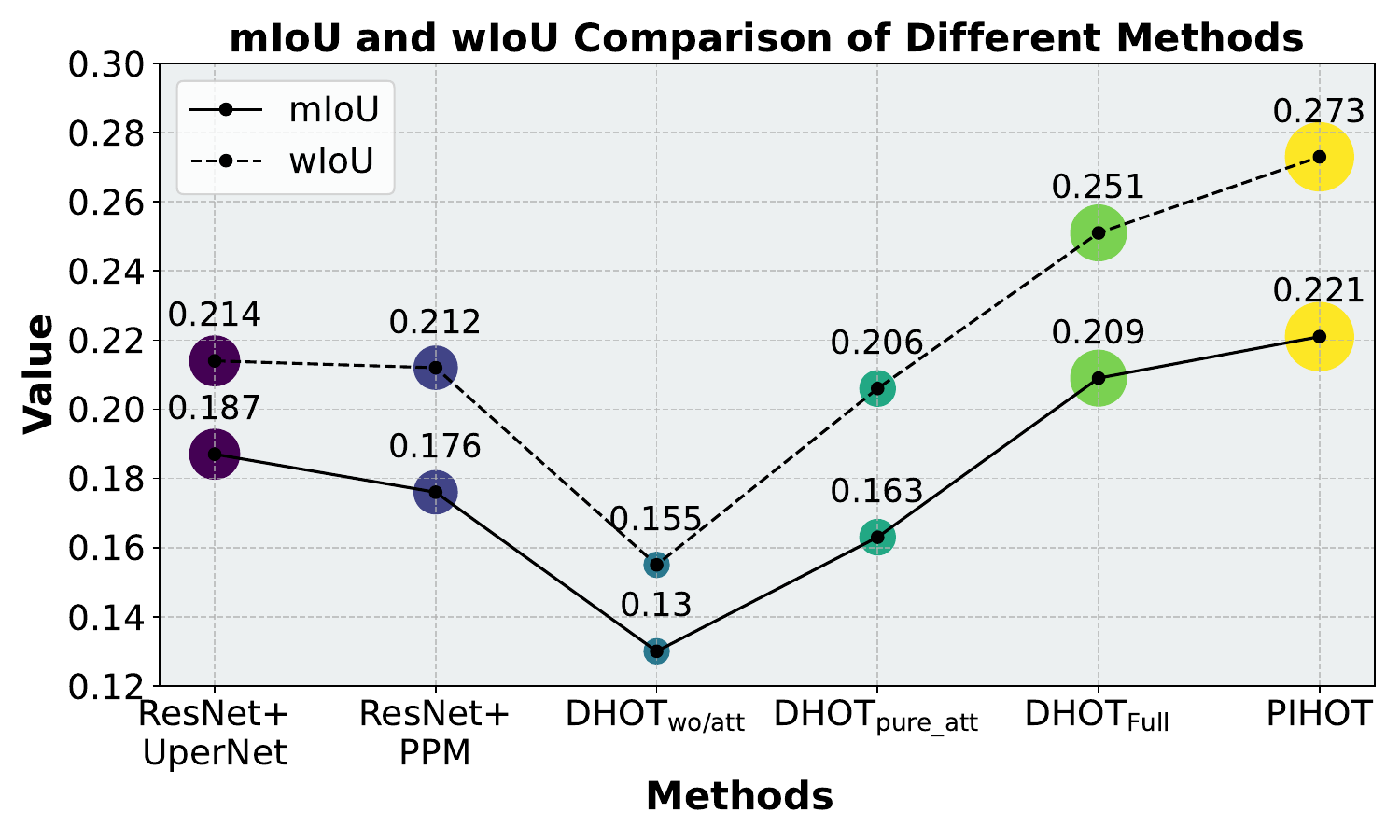}
        \caption{}
    \end{subfigure}
    \caption{(a) compares the SC-Acc. and C-Acc. metrics across different methods. (b) compares the mIoU and wIoU metrics across different methods. The performance of the models is represented by the size of the dots. The proposed PIHOT method outperforms the current HOT model.}
    \label{fig:results_all}
\end{figure}

\begin{figure*}[!h]
\centering
\includegraphics[width=\linewidth]{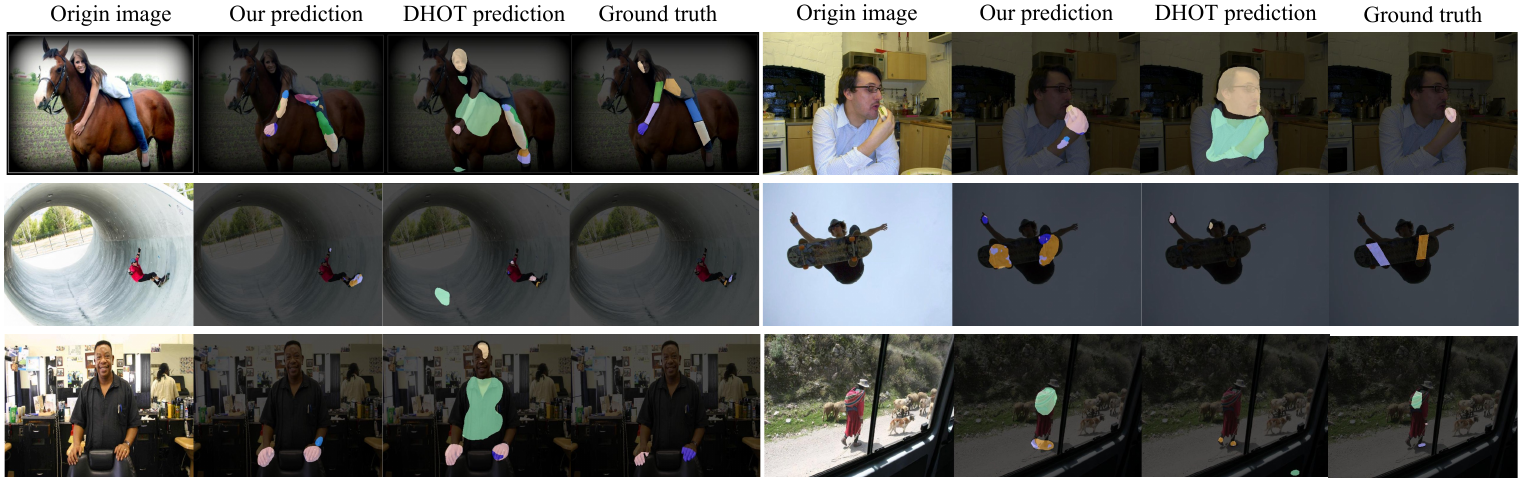}
\caption{Visualizing the segmentation results of HOT on a portion of the dataset. Each group of pictures is separated into four sections: the original image, our method's predicted results, the predicted results from DHOT, and ground truth.}
\label{fig:results_a}
\end{figure*}

\begin{table*}[h]
\centering
\setlength{\tabcolsep}{3.3mm}{
\begin{threeparttable}
\begin{tabular}{cccccc|cccc}
\hline 
\multirow{2}{*}{Row} & \multirow{2}{*}{Baseline} & \multirow{2}{*}{OI} & \multirow{2}{*}{IPI} & \multirow{2}{*}{SPO} & \multirow{2}{*}{IDSI} & \multicolumn{4}{c}{HOT-Annotated}   \\ \cline{7-10} 
                     &                           &                     &                      &                     &                      & SC-Acc. & C-Acc. & mIoU & wIoU \\ \hline
1                    & $\checkmark$              &                     &                      &                     &                      & 40.5      & 73.8     & 0.214   & 0.263   \\
2                    &                           & $\checkmark$        &                      &                     &                      & 42.4(\textcolor{red}{+7\%})      & 74.8(\textcolor{red}{+1\%})     & 0.214(+0\%)   & 0.269(\textcolor{red}{+2\%})   \\
3                    &                           & $\checkmark$        & $\checkmark$         &                     &                      & 44.2(\textcolor{red}{+4\%})      & 79.4(\textcolor{red}{+6\%})     & 0.221(\textcolor{red}{+3\%})   & 0.265(-1\%)   \\
4                    &                           & $\checkmark$        & $\checkmark$         & $\checkmark$        &                      & 44.6(\textcolor{red}{+1\%})      & 79.7(\textcolor{red}{+0.4\%})     & 0.231(\textcolor{red}{+5\%})   & 0.280(\textcolor{red}{+6\%})   \\
\rowcolor[gray]{0.9}5                    &                           & $\checkmark$        & $\checkmark$         & $\checkmark$        & $\checkmark$         & \textbf{45.3}(\textcolor{red}{+2\%})    & \textbf{80.7}(\textcolor{red}{+1\%})  & \textbf{0.236}(\textcolor{red}{+2\%})  & \textbf{0.286}(\textcolor{red}{+2\%})   \\ \hline
\end{tabular}
\begin{tablenotes}
        \footnotesize
        \item OI stands for the object inpainting module, IPI represents the instances perspective interaction mechanism, SPO denotes the space perception operation, and IDSI indicates the instances depth space interaction mechanism. The baseline refers to the simple model without these four modules.
\end{tablenotes}
\end{threeparttable}}
\caption{Ablation study 1: different components on HOT-Annotated dataset.}
\label{tab:ab_hot_a}
\end{table*}

\begin{table*}[h]
\centering
\setlength{\tabcolsep}{3.3mm}{
\begin{tabular}{cccccc|cccc}
\hline
\multirow{2}{*}{Row} & \multirow{2}{*}{Baseline} & \multirow{2}{*}{OI} & \multirow{2}{*}{IPI} & \multirow{2}{*}{SPO} & \multirow{2}{*}{IDSI} & \multicolumn{4}{c}{HOT-Generated}   \\ \cline{7-10} 
                     &                           &                     &                      &                     &                      & SC-Acc. & C-Acc. & mIoU & wIoU \\ \hline
1                    & $\checkmark$              &                     &                      &                     &                      & 27.7      & 57.1     & 0.135   & 0.177   \\
2                    &                           & $\checkmark$        &                      &                     &                      & 30.9(\textcolor{red}{+12\%})      & 67.3(\textcolor{red}{+18\%})     & 0.128(-7\%)   & 0.191(\textcolor{red}{+8\%})   \\
3                    &                           & $\checkmark$        & $\checkmark$         &                     &                      & 31.3(\textcolor{red}{+1\%})      & 71.6(\textcolor{red}{+6\%})     & 0.134(\textcolor{red}{+5\%})   & 0.176(-8\%)   \\
4                    &                           & $\checkmark$        & $\checkmark$         & $\checkmark$        &                      & 34.4(\textcolor{red}{+10\%})      & 71.7(\textcolor{red}{+0.1\%})     & 0.169(\textcolor{red}{+24\%})   & 0.202(\textcolor{red}{+15\%})   \\
\rowcolor[gray]{0.9}5                    &                           & $\checkmark$        & $\checkmark$         & $\checkmark$        & $\checkmark$         & \textbf{34.9}(\textcolor{red}{+1\%})       & \textbf{76.3}(\textcolor{red}{+6\%})     & \textbf{0.169}(+0\%)     & \textbf{0.212}(\textcolor{red}{+5\%})   \\ \hline
\end{tabular}}
\caption{Ablation study 2: different components on HOT-Generated dataset.}
\label{tab:ab_hot_g}
\end{table*}

\subsection{Experimental settings}
\textbf{Datasets}. Three datasets, namely HOT-Annotated, HOT-Generated, and Full Set, are used to evaluate the effectiveness of the proposed method. Specifically, within the HOT-Annotated subset, 5,235 images and 20,273 contact areas are sourced from V-COCO~\cite{gupta2015visual}, 9,522 images and 45,645 contact areas from HAKE~\cite{li2020pastanet}, and 325 images and 1,170 contact areas from the Watch-n-Patch~\cite{wu2015watch}. HOT-Generated originates from data collected from the PROX~\cite{hassan2019resolving} and SMPL-X~\cite{pavlakos2019expressive}. HOT-Generated includes 95,179 contact areas in 20,205 images. The Full Set dataset consists of the union of HOT-Annotated and HOT-Generated.

\textbf{Evaluation metric}. We follow the metrics proposed by Chen et al.~\cite{hot} to evaluate the proposed method, i.e., Semantic contact accuracy (SC-Acc.), Contact accuracy (C-Acc.), Mean IoU (mIoU), and Weighted IoU (wIoU). During inference, the human mask can be generated using existing human segmentation models as an alternative to human annotation masks.

\subsection{Effectiveness for regular HOT}

The performance of different methods on the HOT-Annotated and HOT-Generated datasets is shown in Table~\ref{tab:results-annotated}. For HOT-Annotated, PIHOT surpasses other methods on this dataset, achieving state-of-the-art performance. Specifically, PIHOT outperforms DHOT by 11\%, 10\%, and 10\% on metrics SC-Acc., mIoU, and wIoU, respectively, with a notable improvement of 14\% observed in metric C-Acc.. This indicates that PIHOT is more effective in identifying object contact areas. Additionally, on the HOT-Generated dataset (as shown in the right of Table \ref{tab:results-annotated}), PIHOT achieves performance metrics of 34.9, 76.3, 0.169, and 0.212, outperforming DHOT by 15\%, 41\%, 22\%, and 27\%, respectively. 

For the combined dataset Full Set, the experimental performance is illustrated in Figure \ref{fig:results_all}. Figure \ref{fig:results_all}(a) shows the performance of various methods on SC-Acc. and C-Acc., while Figure \ref{fig:results_all}(b) shows the performance on mIoU and wIoU. We use the size of the dots to indicate performance levels to make it easier for readers to understand. It is evident that PIHOT significantly outperforms the existing HOT model, achieving SOTA performance. Specifically, PIHOT improves the metrics by 16\%, 17\%, 6\%, and 9\% compared to the second-best method, achieving 77.5, 42.4, 0.273, 0.221 in SC-Acc., C-Acc. mIoU and wIoU, respectively.

\subsection{Ablation study}
Table \ref{tab:ab_hot_a} illustrates the impact of different components on the performance of the proposed method. For the HOT-Annotated dataset, when using the baseline model, the results are only 40.5, 73.8, 0.214, and 0.263. When the OI module is added, there is an improvement of 7\%, 1\%, 0\%, and 2\% in the four metrics, respectively. Further performance enhancement is observed with the addition of IPI, resulting in 4\%, 6\%, and 3\% improvements in SC-ACC., C-ACC., and mIoU, respectively. However, there is a decrease of 1\% in wIoU. The influence of incorporating SPO module on the experimental results is demonstrated in the fourth row of Table \ref{tab:ab_hot_a}, showing an improvement of 1\%, 0.4\%, 5\%, and 6\%. Upon integrating the IDSI mechanism, the results reach optimum values at 45.3, 80.7, 0.236, and 0.286. 

Similarly, results on the HOT-Generated are listed in Table \ref{tab:ab_hot_g}. Extensive experiments indicate that the optimal performance is achieved when OI, IPI, OI, and IDSI are simultaneously utilized, resulting in 34.9, 76.3, 0.169, and 0.212 on HOT-Generated. The performances of different components on the Full Set are written in supporting materials.

Before finalizing the Inpainting Model and Depth Map model, we conducted experiments with other similar models and ultimately selected the LaMa and ZoeDepth models~\cite{suvorov2022resolution,bhat2023zoedepth}, which achieved state-of-the-art performance. Other models (as shown in Table \ref{tab:as_othermodel}), such as MAE and MDENet~\cite{he2022masked}, demonstrated performance scores of 43.4, 75.6, 0.279, and 0.265 on HOT-Annotation, falling short of the selected models by 4\%, 6\%, 6\%, and 2\% across the four metrics, respectively.

\begin{table}[]
\begin{tabular}{c|cccc}
\hline
Method        & SC-Acc. & C-Acc. & mIoU  & wIoU  \\ \hline
MAE+MDENet    & 43.4    & 75.6   & 0.279 & 0.265 \\
Lama+ZoeDepth & 45.3    & 80.7   & 0.236 & 0.286 \\ \hline
\end{tabular}
\caption{The performance of different depth models and restoration models on HOT-Annotated dataset.}
\label{tab:as_othermodel}
\end{table}

\subsection{Visualization}
To visually compare the performance of the proposed method against the SOTA method, we conducted visualizations on some images from the dataset. Figure \ref{fig:results_a} respectively demonstrate the result images on the HOT dataset. The proposed method accurately segments the contact area between the human and the object, even in cases where the object is partially hidden from view. In the same way, our approach surpasses the most recent DHOT method in segmentation effectiveness.

\subsection{Limitations and future directions}
PIHOT may be limited to extreme occlusion scenarios or dynamic scenes. Future directions may lie on a unified multi-task design to jointly optimize human prediction, depth estimation, and HOT detection, or a novel HOT detection framework for 3D real-time video streams.

\section{Conclusion}
This study introduces a technique called PIHOT that aims to solve the problem of inaccurate segmentation in the HOT task caused by occlusion between humans and objects. Initially, PIHOT uses human masks to identify the areas where humans are present and then restores the obstructed parts of the objects. Second, it uses the depth map to capture the spatial relationship between humans and objects, allowing the method to have a clear understanding of their spatial positioning. Moreover, IPI and IDSI mechanisms are developed to emphasize crucial contact areas between humans and objects by incorporating object information and spatial position features. Extensive experiments have shown that PIHOT outperforms all other methods on the benchmark datasets for HOT detection tasks. The concept of HOT was first introduced by ~\cite{hot}, and it has since become a vibrant and rapidly growing research area. Our research has injected significant momentum into the development of this field.

\section{Acknowledgments}
This work was supported in part by the National Natural Science Foundation of China under Grant 62202174, in part by the Basic and Applied Basic Research Foundation of Guangzhou under Grant 2023A04J1674, in part by The Taihu Lake Innocation Fund for the School of Future Technology of South China University of Technology under Grant 2024B105611004, and in part by Guangdong Science and Technology Department  Grant 2024A1313010012.

\bibliography{ref}

\appendix
\clearpage
% \twocolumn  % 切换为双栏布局

\section{Appendix}  % 手动添加附录标题，不带编号

\subsection{Instances Perspective Interaction Mechanism}
\begin{figure}[h]
\centering %表示居中
\includegraphics[width=\linewidth]{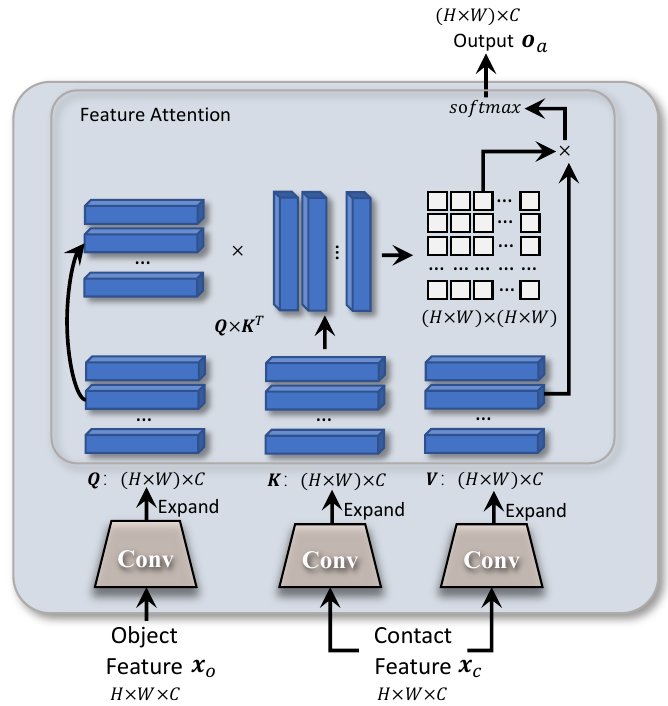}
% [height=4.5cm]表示高度
%[width=9.5cm]表示宽度
%{111.eps}表示eps格式的图片，名为111
\caption{The structure of the IPI mechanism. Using the object feature $\bm{x}_o$ and the contact feature $\bm{x}_i$ as input, the model employs the attention mechanism to identify key feature parts in the original image.}
%图片的名称
\label{fig:ofa}
%图片的标签，用于文章中的引用，注意到标签的数字与实际文章显示的数字可能不同
% \vspace{-0.5cm}
\end{figure}

The instances perspective interaction (IPI) mechanism is shown in Figure \ref{fig:ofa}. The feature $\bm{x}_o$ is first expanded into a 2D matrix using a convolutional layer, resulting in matrix $\bm{Q}$. Simultaneously, the feature $\bm{x}_c$ is passed through two distinct convolutional layers to produce matrices $\bm{K}$ and $\bm{V}$. The significance of each position, represented by the weights, is determined by multiplying matrix $\bm{Q}$ with the transpose of matrix $\bm{K}$. Finally, these weights are multiplied by matrix $\bm{V}$, yielding the output feature matrix $\bm{o}_a$.

\subsection{Instances Depth Space Interaction}
\begin{figure}[h]
\centering %表示居中
\includegraphics[width=\linewidth]{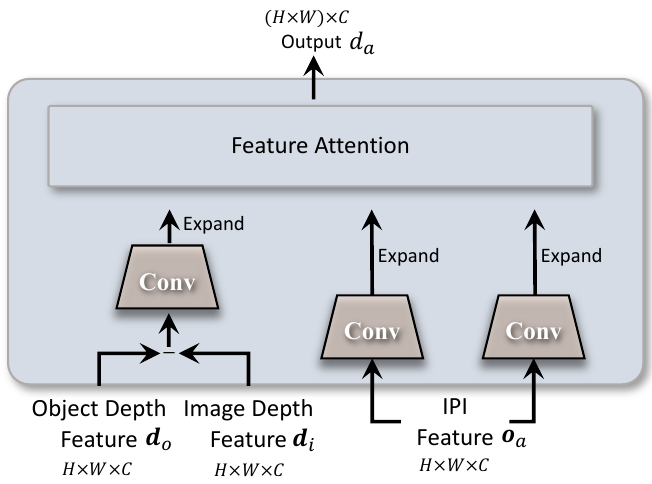}
%[width=9.5cm]表示宽度
%{111.eps}表示eps格式的图片，名为111
\caption{The structure of the IDSI mechanism. The depth map information $\bm{d}_s$ is obtained from Eqs. 5 and 6 of main text. The aim is to direct the model to concentrate on specific areas by considering the spatial relationship between humans and objects. IPI stands for instances perspective interaction module.}
%图片的名称
\label{fig:dfa}
%图片的标签，用于文章中的引用，注意到标签的数字与实际文章显示的数字可能不同
% \vspace{-0.5cm}
\end{figure}

The Instances Depth Space Interaction (IDSI) mechanism is shown in Figure \ref{fig:dfa}. The IDSI mechanism receives three inputs: $\bm{d}_o$, $\bm{d}_i$, and the output of the IPI module, $\bm{o}_a$. $\bm{d}_o$ and $\bm{d}_i$ are subtracted and then normalized to obtain $\bm{d}_s$ (as described in the IDSI subsection in the main text). $\bm{d}_s$ is subsequently passed through a convolutional layer and flattened into a 2D vector. Meanwhile, $\bm{d}_o$ is fed into two different convolutional modules to extract distinct features. Finally, \texttt{Feature Attention} is applied for further processing.

\subsection{Parameter settings}
The proposed method utilizes ResNet-50 as the backbone for feature extraction. The parameters $\alpha$ and $\beta$ are both set to 0.1. To mitigate the influence of excessive background categories on the experiments, we set the ratio of classification loss for the background to 0.2. The network is optimized using the Adam optimizer with a learning rate of 1e-5. All experiments are conducted on Ubuntu 20.04 with 8 NVIDIA A6000 GPUs (48GB). PyTorch version 1.11.0 and torchvision version 0.12.0 are used, with a batch size of 4 per GPU. During training, random data augmentation techniques are applied, including random flipping, random cropping, and so on.
\begin{table*}[ht]
\centering
\begin{tabular}{cccccc|cccc}
\hline
\multirow{2}{*}{Row} & \multirow{2}{*}{Baseline} & \multirow{2}{*}{OI} & \multirow{2}{*}{IPI} & \multirow{2}{*}{SPO} & \multirow{2}{*}{IDSI} & \multicolumn{4}{c}{Full Set}   \\ \cline{7-10} 
                     &                           &                     &                      &                     &                      & SC-Acc. & C-Acc. & mIoU & wIoU \\ \hline
1                    & $\checkmark$              &                     &                      &                     &                      & 31.8      & 66.5     & 0.180   & 0.194   \\
2                    &                           & $\checkmark$        &                      &                     &                      & 38.4(\textcolor{red}{+21\%})      & 71.3(\textcolor{red}{+7\%})     & 0.209(\textcolor{red}{+16\%})   & 0.263(\textcolor{red}{+36\%})   \\
3                    &                           & $\checkmark$        & $\checkmark$         &                     &                      & 40.7(\textcolor{red}{+6\%})      & 75.4(\textcolor{red}{+6\%})     & 0.211(\textcolor{red}{+1\%})   & 0.261(-1\%)   \\
4                    &                           & $\checkmark$        & $\checkmark$         & $\checkmark$        &                      & 42.0(\textcolor{red}{+3\%})      & 76.7(\textcolor{red}{+2\%})     & 0.219(\textcolor{red}{+4\%})   & 0.270(\textcolor{red}{+3\%})   \\
\rowcolor[gray]{0.9}5                    &                           & $\checkmark$        & $\checkmark$         & $\checkmark$        & $\checkmark$         & \textbf{42.4}(\textcolor{red}{+1\%})       & \textbf{77.5}(\textcolor{red}{+1\%})     & \textbf{0.221}(\textcolor{red}{+1\%})     & \textbf{0.273}(\textcolor{red}{+1\%})   \\ \hline
\end{tabular}
\caption{Ablation study 3: different components on Full Set on dataset.}
\label{tab:ab_hot_all}
\end{table*}

% \begin{table*}[ht]
% \centering
% % \vspace{-0.1cm}
% \begin{tabular}{c|c|c|cccc}
% \hline
% Model & Time(ms) & Epoch & SC-Acc.                                                & C-Acc.                                                 & mIoU                                                    & wIoU                                                    \\ \hline
% DHOT  & \textbf{181}   & 14    & 40.7                                                   & 70.7                                                   & 0.215                                                   & 0.260                                                   \\
% Ours  & 208   & \textbf{12}     & \begin{tabular}[c]{@{}c@{}}\textbf{45.3}\\ (\textcolor{red}{+11\%})\end{tabular} & \begin{tabular}[c]{@{}c@{}}\textbf{80.7}\\ (\textcolor{red}{+14\%})\end{tabular} & \begin{tabular}[c]{@{}c@{}}\textbf{0.236}\\ (\textcolor{red}{+10\%})\end{tabular} & \begin{tabular}[c]{@{}c@{}}\textbf{0.286}\\ (\textcolor{red}{+10\%})\end{tabular} \\ \hline
% \end{tabular}
% \caption{Comparison of efficiency with the latest methods.}
% \label{tab:table1}
% % \vspace{-0.2cm}
% \end{table*}

\subsection{Ablation study}
Table \ref{tab:ab_hot_all} illustrates the impact of different components on the performance of the proposed method in Full Set datasets. When the OI, IPI, SPO, and IDSI modules were all removed, the model's performance was 31.8, 66.5, 0.180, and 0.194 on the SC-Acc., C-Acc., mIoU, and wIoU metrics, respectively. After adding the OI module, the model's performance on these four metrics improved by 21\%, 7\%, 16\%, and 36\%, reaching 38.4, 71.3, 0.209, and 0.263, respectively. With the IPI module added, the SC-Acc., C-Acc., and mIoU metrics improved by 6\%, 6\%, and 1\%, respectively, although the wIoU metric decreased by 1\%. After incorporating the SPO module, the model's performance reached 42.0, 76.7, 0.219, and 0.270. Finally, with all modules applied, the model's performance on the Full Set dataset achieved 42.4, 77.5, 0.221, and 0.273.

% When all four components are utilized, the PIHOT model achieves results of 42.4, 77.5, 0.221, and 0.273 on the Full Set, respectively.

\subsection{The role of depth map $\bm{d}_s$}
\begin{figure}[h]
\centering %表示居中
\includegraphics[width=\linewidth]{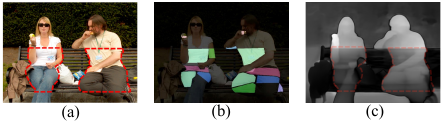}
\vspace{-0.7cm}
%[width=9.5cm]表示宽度
%{111.eps}表示eps格式的图片，名为111
\caption{Description of contact area occlusion problem.}
%图片的名称
\label{fig:rebuttal_fig1}
%图片的标签，用于文章中的引用，注意到标签的数字与实际文章显示的数字可能不同
\vspace{-0.2cm}
\end{figure}

Ideally, subtracting the depth maps ($\bm{d}_s$) would leave only the parts of the human and the occluded object. However, since the disparity of the resulting depth maps between the repaired image and the original image from the model, it turns out that the body is brighter, the background and object parts are darker, and occlusions have intermediate brightness, as seen in Figure \ref{fig:rebuttal_fig1}(c). Hence, the boundary of occlusion becomes more obvious, that's our motivation to segment occlusive and occlusion-free parts.

% \section{Performance of proposed method}
% We have conducted experiment on efficiency in Tab. \ref{tab:table1}. Although our method is 27ms slower than the SOTA method  DHOT, it converges quickly, using 12 epochs. On the HOT-Annotated, our method outperforms DHOT by 11\%, 14\%, 10\%, and 10\% on four metrics, demonstrating improved effectiveness despite the slight increase in processing time.

\end{document}